\documentclass[sigconf]{acmart}
\setcopyright{none}

\AtBeginDocument{%
  \providecommand\BibTeX{{%
    \normalfont B\kern-0.5em{\scshape i\kern-0.25em b}\kern-0.8em\TeX}}}

\acmConference[KDD '23]{In Proceedings of the 29th ACM SIGKDD Conference on Knowledge Discovery and Data Mining}{August 6-10, 2023}{Long Beach, CA}
\begin{document}

\title{Mitigating Bias in Conversations: A Hate Speech Classifier and Debiaser with Prompts}

\author{Shaina Raza}
\affiliation{%
  \institution{Vector Institute of Artificial Intelligence}
  \city{Toronto}
  \state{ON}
  \country{Canada}
 }
 \email{shaina.raza@vectorintitute.ai}

\author{Chen Ding}
\affiliation{%
  \institution{Toronto Metropolitan University}
  \city{Toronto}
  \state{ON}
  \country{Canada}}
  \email{cding@torontomu.ca}

\author{Deval Pandya}
\affiliation{%
  \institution{Vector Institute of Artificial Intelligence}
  \city{Toronto}
  \state{ON}
  \country{Canada}
}
\email{deval.pandya@vectorinstitute.ai}

\renewcommand{\shortauthors}{SRaza, et al.}

\begin{abstract}
Discriminatory language and biases are often present in hate speech during conversations, which usually lead to negative impacts on targeted groups such as those based on race, gender, and religion. To tackle this issue, we propose an approach that involves a two-step process: first, detecting hate speech using a classifier, and then utilizing a debiasing component that generates less biased or unbiased alternatives through prompts. We evaluated our approach on a benchmark dataset and observed reduction in negativity due to hate speech comments. The proposed method contributes to the ongoing efforts to reduce biases in online discourse and promote a more inclusive and fair environment for communication.
\end{abstract}




\maketitle

\section{Introduction}
In the era of social media and online platforms, communication and idea exchange have reached at its peak. Despite many benefits, these platforms also facilitate the spread of hate speech and offensive language. Hate speech often contains biases, perpetuating stereotypes and discriminatory language, which exacerbates the negative impact of such content on different targeted groups (based on race, gender, religion) \cite{Garg2023}. Addressing these biases is a crucial step towards developing unbiased text processing systems and fostering healthy online interactions.

In this paper, we propose a debiasing technique that leverages language generation and in-context prompting \cite{liu2021pre} to minimize the influence of lexical biases. A prompt is an instruction usually consisting of a few words or sentences that provide context or constraints for the model to follow . The method works by first detecting hate speech using a classifier, then employing a debiasing component that generates less biased alternatives through incorporating context-aware prompts designed to reduce the presence of biased language patterns.  

We evaluate our approach on benchmark dataset and demonstrate its effectiveness in debiasing hate speech texts. The results show a classifier accuracy of 95\% and debiasing accuracy of 89\%, along with a notable reduction in negative sentiment within hate speech comments. This method contributes to the ongoing efforts to reduce biases in online discourses.

\section{Related Work}
\label{Related Work}
Bias in language models and embeddings is a broad and subjective topic \cite{raza2022dbias} . Research has identified gender bias in popular embeddings such as GloVe and Word2Vec \cite{bolukbasi2016man} and quantified biases using the word embedding association test (WEAT) \cite{caliskan2017semantics}
\cite{may2019measuring}. Efforts have been made to reduce biases in Transformer-based language models like BERT and GPT-3 \cite{bartl2020unmasking} and in conversational AI systems \cite{barikeri2021redditbias}.

Hate speech refers to the use of derogatory, abusive or threatening language towards individuals or groups based on their race, ethnicity, gender, religion, sexual orientation or any other characteristic.  Several studies have focused on developing machine learning models for detecting hate speech in text \cite{davidson2017automated}.   One work \cite{smith-etal-2022-im} presents a dataset, HOLISTIC BIAS, which consists of nearly 600 descriptor terms across 13 different demographic axes. The work demonstrates that this dataset is highly effective for measuring previously unmeasurable biases in token likelihoods and generations from language models, as well as in an offensiveness classifier.  

Another work \cite{huang2019reducing} quantifies sentiment bias through  individual and group fairness metrics \cite{raza2022dbias}  and proposes embedding and sentiment prediction-derived regularization on the language model’s latent representations. The role of individual neurons and attention heads in mediating gender bias across three datasets designed to gauge a model’s sensitivity to gender bias are also studied \cite{vig2020investigating}.

A related paper \cite{liu2021mitigating} describes metrics for measuring political bias in language generation and proposes a reinforcement learning framework for mitigating political biases in generated text.  StereoSet \cite{nadeem2020stereoset}, a large-scale natural English dataset to measure stereotypical biases in four domains: gender, profession, race, and religion, is presented and it is shown that popular models like BERT, GPT-2, RoBERTa, and XLnet exhibit strong stereotypical biases. 

A novel approach to mitigate gender disparity in text generation by learning a fair model during knowledge distillation is presented in a study \cite{gupta2022mitigating} and two modifications based on counterfactual role reversal are proposed—modifying teacher probabilities and augmenting the training set. A related work \cite{abid2021persistent} shows that GPT-3, a state-of-the-art contextual language model, captures persistent Muslim-violence bias, demonstrating that it appears consistently and creatively in different uses of the model. These biases become even severe compared to biases about other religious groups. 

Prompt engineering has recently emerged as a promising approach to mitigate biases in language models \cite{liu2021pre} . In the context of hate speech detection, few-shot learning \cite{anand2021few} can be useful in scenarios where there is limited labeled data available for a particular language or dialect. Recent studies \cite{anand2021few} have shown promising results in using few-shot learning for hate speech detection, which is also a motivation. In this work, we use the OPT based model for debiasing but we introduce fairness aware prompts to achieve the goal of debiasing the texts.

\section{Methodology}

\subsection{Hate speech classifier}
We utilize the BERT \cite{devlin2018bert}  for building an efficient hate speech detection. The input to the BERT encoder consists of tokenized text sequences with special tokens [CLS] and [SEP] for classification and separation. The output from the BERT encoder is passed through a SoftMax layer to provide a probability distribution for the text being classified as hate speech or non-hate speech. The hate speech classifier is trained on a labeled dataset using binary cross-entropy loss as the objective function, which aims to minimize the difference between the predicted probability distribution and the true binary labels, encouraging the model to accurately classify hate speech and non-hate speech comments.

\subsection{Debiasing model}
We use the pre-trained OPT (OpenAI's Pre-trained Transformer) \cite{radford2018improving} model for debiasing hate speech classification. We employ the few-shot learning with prompts to further refine the debiasing process. Our approach to debiasing through prompts is inspired by similar works such as \cite{wu2022leveraging} , however, we make our own changes through task-specific examples.

To incorporate the debiasing prompts into the OPT model, we provide the prompt,  the input text, resulting in a new input sequence. The OPT model processes this combined sequence and generates the debiased output based on the contextualized embeddings of both the prompt and the input text.  We adjust the temperature of the OPT model during inference to control the level of randomness and diversity in the generated text samples. This helps to mitigate biases while preserving the overall meaning of the input text.

\subsection{Pipeline}
Our pipeline stacks both the hate speech classification and debiasing models as follows: (i) pass the input text through the hate speech classifier to obtain the predicted probability of it being hateful or non-hateful, and (ii) pass the hateful text through the  debiasing model using few-shot learning with prompts to generate the debiased text during the language generation task. This two-stage approach first classifies the hate content in the input text, and if it is deemed hateful by the model, then it debiases it.  We show our pipeline approach in Figure 1.
\begin{figure}[h]
  \centering
  \includegraphics[width=0.9\columnwidth]{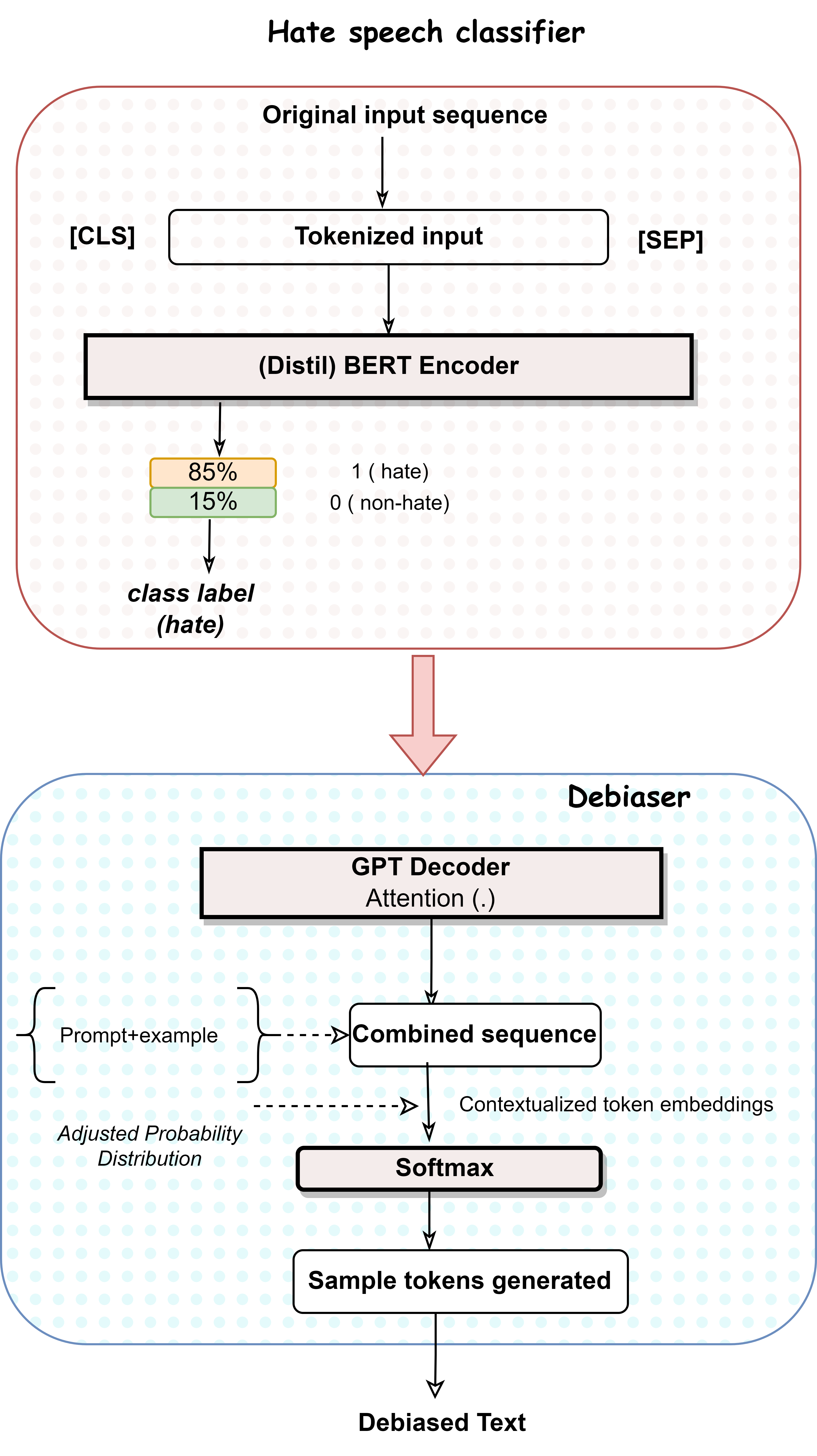}
  \caption{Proposed framework consisting of a classifier for hate speech and debiaser with prompt and examples.}
  \label{fig:my_figure}
\end{figure}
\section{Experiment and Results}
\subsection{Dataset}
In this study, we utilize the Hate Speech Dataset from a white supremacy online community \cite{grimminger-klinger-2020-hate} . The dataset consists of a diverse range of text samples, including both overt and subtle instances of hate speech, posing significant challenges for automated classification and debiasing models.  The dataset contains a total of 10,568 sentences, classified as conveying hate speech or not. The size of dataset, along with the average sentence length, word count, and vocabulary size for each class. 
\begin{itemize}
    \item HATE:   Sentences: 1,119,  Sentence length: 20.39 ± 9.46, Word count: 24,867, Vocabulary: 4,148.
    \item     NOHATE:  Sentences: 8,537,  Sentence length: 15.15 ± 9.16, - Word count: 144,353, - Vocabulary: 13,154.
\end{itemize}

In this study, we address the class imbalance problem in our dataset by using a hybrid approach combining under-sampling and over-sampling. We start by randomly removing instances from the 'NOHATE' class to balance the representation. After this, we augment the 'HATE' class byduplicating instances using a method like SMOTE. This ensures both classes are equally represented, improving our model's ability to learn from both hate speech and non-hate speech instances. 
\subsection{Hyperparameters and Evaluation}
We fine-tuned the BERT model for hate speech classification to develop an efficient and accurate classifier. For the debiaser model, we use the GPT-2 -small model with 117M parameters. We tuned the model temperature (0.1 - 1.0) in our debiaser model to balance diversity in the generated text samples \cite{gptwork}. Additionally, we utilized few-shot learning with prompts to further refine the debiasing process. We used 5 and 10 examples per category in our few-shot learning experiments. For the classification task and to train the whole pipeline, we used 3 epochs, optimizing the parameters using the Adam optimizer with a learning rate of 5e-5, weight decay of 0.5 and an epsilon value of 1e-8.  We searched a grid of hyperparameters, including batch sizes of 4, 8, and 16, 32, 64 . We limited the input sequences to 128 (subword) tokens and trained the model in batches of 16 (to avoid out-of-memory issues). We employ F1 score (by calculating the harmonic mean precision and recall) and accuracy. For training, we used Google Colab Pro, which provided access to an NVIDIA Tesla T4 GPU with 16 GB of memory. 

\subsection{Performance Evaluation}
\begin{table}[t]
    \centering
    \caption{Performance comparison of different methods on the hate speech classification and debiasing tasks.}
     \begin{tabular}{l c c}
    \toprule
    \textbf{Method} & \textbf{F1 Score $\pm$ SD} \\
    \midrule
    \multicolumn{2}{c}{\textbf{Classifier performance}} \\
    Rule-based & 0.60 $\pm$ 0.03 \\
    SVM & 0.70 $\pm$ 0.02 \\
    BERT-d & 0.90 $\pm$ 0.04 \\
    RoBERTa-HS & 0.92 $\pm$ 0.01 \\
    Proposed & 0.95 $\pm$ 0.05 \\
    \multicolumn{2}{c}{\textbf{Debiasing performance}} \\
    Debiased with zero-shot learning & 0.86 $\pm$ 0.02 \\
    Debiased with few-shot learning (5) & 0.88 $\pm$ 0.03 \\
    Debiased with few-shot learning (10) & 0.89 $\pm$ 0.02 \\
    \bottomrule
    \end{tabular}
    \label{tab:performance_comparison}
\end{table}

Table 1 presents the results of our proposed method for mitigating biases in hate speech classification. The classifier's performance is evaluated using F1-score metrics. Further, we introduce a novel measure - the bias score - to assess the effectiveness of our debiasing model.

In this study, our classifier generates a quantifiable bias score for each text, reflecting the degree of hate speech bias. Each text is scored both before and after the debiasing process. A decrease in this bias score post-debiasing is indicative of successful bias mitigation. This metric provides us a way to numerically gauge the effectiveness of our proposed debiasing model.

The results shows that the proposed method achieves the highest F1 score of about 95\%, outperforming all other classification methods. Moreover, the debiasing performance of the proposed method with few-shot learning and prompts achieves an impressive F1 score of 89\%, with a low standard deviation. This indicates that the proposed method can effectively mitigate biases in hate speech classification.

  To evaluate the performance of the debiaser model, we conducted experiments with zero-shot and then 5 and 10 prompts with examples, respectively. The results show that using few-shot learning with prompts improves the F1 score compared to zero-shot learning. This indicates that the model is better able to mitigate biases in the input text with the guidance of a few examples. This technique has the potential to enhance the performance of the debiasing model further with more diverse and relevant prompt examples, which could be explored in future work.
  
\textbf{Classification performance}: Next, we show the performance of the classifier model on a sample of 100 examples and report the results in Figure 2.
\begin{figure}[h]
  \centering
  \includegraphics[width=0.5\textwidth]{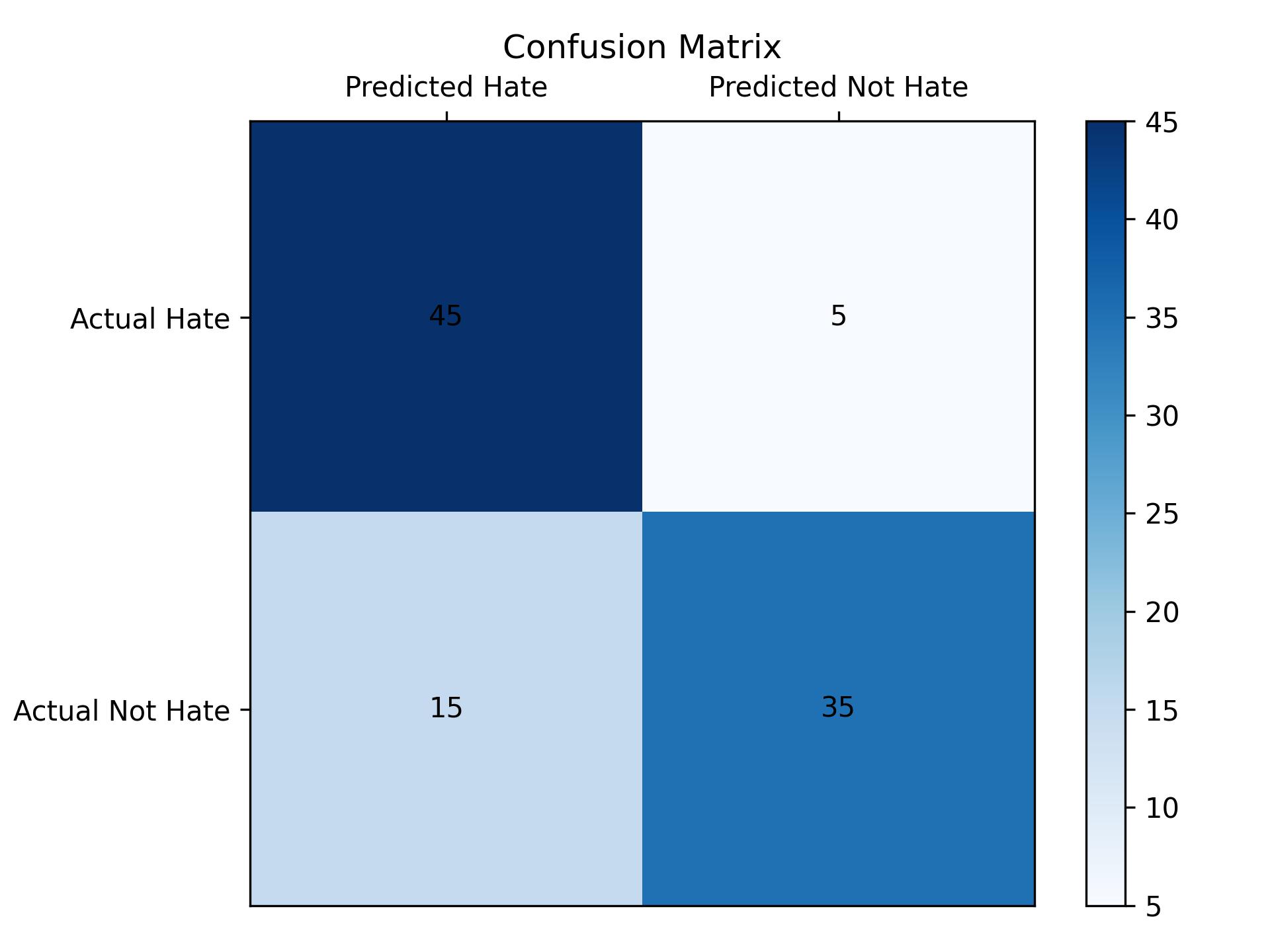}
  \caption{Confusion matrix for hate speech detection.}
  \label{fig:confusion_matrix}
\end{figure}
We observe in the Figure 2, the model has made 45 true positive predictions, 5 false positive predictions, and 35 true negative predictions. The false negative count is 15.  The high true positive rate indicates that the model is correctly identifying a majority of the hate speech sentences, while the low false positive rate indicates that the model is not incorrectly labeling non-hate speech sentences as hate speech. However, the false negative count is 15, indicating that there is still room for improvement in correctly identifying all instances of hate speech.  Since, these results are based on a sample of 100 instances and may not be representative of the model's performance on a larger dataset. Further evaluation and tuning of the model may be necessary to improve its overall performance.

\textbf{Debiasing Performance}: To further measure the effectiveness of our debiasing method, we measure and compare a range of performance metrics before and after debiasing. We first train our classifier on the original, biased dataset, and calculate the bias score. We then apply our debiasing process to the data, train the classifier on the debiased dataset, and again calculate the bias score. The change in the bias score serves as an indication of the effectiveness of our debiasing method.

As shown in Table 2, our debiasing method successfully reduced the bias score by 30\%, from 65\% to 35\%. This indicates a significant reduction in the model's inclination to misclassify certain non-hateful speech as hateful. However, these improvements were accompanied by a slight decrease in hate speech detection accuracy, with the accuracy dropping from 85\% to 83\%. Additionally, there was a slight increase in the false negatives rate, which rose from 15\% to 17\%. These findings suggest that the debiasing caused slightly less accuracy in identifying the hateful speech. On the other hand, the false positives rate decreased from 20\% to 12\%, indicating an improvement in correctly identifying non-hateful speech .
\begin{table}[t]
\caption{Performance comparison of original and debiased data on the hate speech classification task.}
\begin{tabular}{p{3.5cm} p{2cm} p{2cm}}
\toprule
\textbf{Metrics} & \textbf{Original Data} & \textbf{Debiased Data} \\
\midrule
\multicolumn{3}{l}{\textbf{Classifier performance}} \\
Hate Speech Detection Accuracy & 85\% & 83\% \\
False Positives Rate & 20\% & 12\% \\
False Negatives Rate & 15\% & 17\% \\
\multicolumn{3}{l}{\textbf{Debiasing performance}} \\
Bias Score* & 65\% & 35\% \\
Reduction in Bias Score & - & 30\% \\
\bottomrule
\end{tabular}
\small{*Bias score represents the degree of bias in the model, with a higher percentage indicating more bias. In this case, it's a measure of how much the model is biased towards incorrectly classifying certain non-hateful speech as hate speech.}
\label{table:2}
\end{table}
\section{Conclusion}
We propose a hate speech classifier and debiaser that employs  prompts to generate less biased alternatives. Our approach first detects hate speech using a classifier and then utilizes a debiasing component that generates less or no biased alternatives.   Our proposed approach has some limitations. One limitation is the size and quality of the available training data, which may not be sufficient to cover all possible scenarios. Another limitation is the need for fine-tuning of the language model, which may require additional resources and expertise. Furthermore, the effectiveness of our approach may be affected by the quality of the underlying language model used for generating alternative text. Nonetheless, further research is needed to address the complex issue of online hate speech and biased language comprehensively.

  \bibliographystyle{ACM-Reference-Format}
  \bibliography{paper}


\begin{thebibliography}{22}


\ifx \showCODEN    \undefined \def \showCODEN     #1{\unskip}     \fi
\ifx \showDOI      \undefined \def \showDOI       #1{#1}\fi
\ifx \showISBNx    \undefined \def \showISBNx     #1{\unskip}     \fi
\ifx \showISBNxiii \undefined \def \showISBNxiii  #1{\unskip}     \fi
\ifx \showISSN     \undefined \def \showISSN      #1{\unskip}     \fi
\ifx \showLCCN     \undefined \def \showLCCN      #1{\unskip}     \fi
\ifx \shownote     \undefined \def \shownote      #1{#1}          \fi
\ifx \showarticletitle \undefined \def \showarticletitle #1{#1}   \fi
\ifx \showURL      \undefined \def \showURL       {\relax}        \fi
\providecommand\bibfield[2]{#2}
\providecommand\bibinfo[2]{#2}
\providecommand\natexlab[1]{#1}
\providecommand\showeprint[2][]{arXiv:#2}

\bibitem[Abid et~al\mbox{.}(2021)]%
        {abid2021persistent}
\bibfield{author}{\bibinfo{person}{Abubakar Abid}, \bibinfo{person}{Maheen
  Farooqi}, {and} \bibinfo{person}{James Zou}.}
  \bibinfo{year}{2021}\natexlab{}.
\newblock \showarticletitle{Persistent anti-muslim bias in large language
  models}. In \bibinfo{booktitle}{\emph{Proceedings of the 2021 AAAI/ACM
  Conference on AI, Ethics, and Society}}. \bibinfo{pages}{298--306}.
\newblock


\bibitem[Anand et~al\mbox{.}(2021)]%
        {anand2021few}
\bibfield{author}{\bibinfo{person}{Aishwarya Anand}, \bibinfo{person}{Murthy
  Devarakonda}, \bibinfo{person}{Manik Gambhir}, \bibinfo{person}{Arnav
  Wadhwa}, {and} \bibinfo{person}{Mark~J Carman}.}
  \bibinfo{year}{2021}\natexlab{}.
\newblock \showarticletitle{Few-Shot Learning for Hate Speech Detection on
  Social Media}. In \bibinfo{booktitle}{\emph{Proceedings of the 1st Workshop
  on Online Abuse and Harms (WOAH 2021)}}. \bibinfo{pages}{30--35}.
\newblock


\bibitem[Barikeri et~al\mbox{.}(2021)]%
        {barikeri2021redditbias}
\bibfield{author}{\bibinfo{person}{Soumya Barikeri}, \bibinfo{person}{Anne
  Lauscher}, \bibinfo{person}{Ivan Vuli{\'c}}, {and} \bibinfo{person}{Goran
  Glava{\v{s}}}.} \bibinfo{year}{2021}\natexlab{}.
\newblock \showarticletitle{RedditBias: A real-world resource for bias
  evaluation and debiasing of conversational language models}.
\newblock \bibinfo{journal}{\emph{arXiv preprint arXiv:2106.03521}}
  (\bibinfo{year}{2021}).
\newblock


\bibitem[Bartl et~al\mbox{.}(2020)]%
        {bartl2020unmasking}
\bibfield{author}{\bibinfo{person}{Marion Bartl}, \bibinfo{person}{Malvina
  Nissim}, {and} \bibinfo{person}{Albert Gatt}.}
  \bibinfo{year}{2020}\natexlab{}.
\newblock \showarticletitle{{Unmasking Contextual Stereotypes: Measuring and
  Mitigating BERT's Gender Bias}}.
\newblock \bibinfo{journal}{\emph{arXiv preprint arXiv:2010.14534}}
  (\bibinfo{year}{2020}).
\newblock
\showeprint[arxiv]{2010.14534}
\urldef\tempurl%
\url{http://arxiv.org/abs/2010.14534}
\showURL{%
\tempurl}


\bibitem[Bolukbasi et~al\mbox{.}(2016)]%
        {bolukbasi2016man}
\bibfield{author}{\bibinfo{person}{Tolga Bolukbasi}, \bibinfo{person}{Kai-Wei
  Chang}, \bibinfo{person}{James~Y Zou}, \bibinfo{person}{Venkatesh Saligrama},
  {and} \bibinfo{person}{Adam~T Kalai}.} \bibinfo{year}{2016}\natexlab{}.
\newblock \showarticletitle{Man is to computer programmer as woman is to
  homemaker? debiasing word embeddings}.
\newblock \bibinfo{journal}{\emph{Advances in neural information processing
  systems}}  \bibinfo{volume}{29} (\bibinfo{year}{2016}).
\newblock


\bibitem[Caliskan et~al\mbox{.}(2017)]%
        {caliskan2017semantics}
\bibfield{author}{\bibinfo{person}{Aylin Caliskan}, \bibinfo{person}{Joanna~J
  Bryson}, {and} \bibinfo{person}{Arvind Narayanan}.}
  \bibinfo{year}{2017}\natexlab{}.
\newblock \showarticletitle{Semantics derived automatically from language
  corpora contain human-like biases}. In \bibinfo{booktitle}{\emph{Proceedings
  of the 2017 Conference on Empirical Methods in Natural Language Processing}}.
  \bibinfo{pages}{2075--2086}.
\newblock


\bibitem[Davidson et~al\mbox{.}(2017)]%
        {davidson2017automated}
\bibfield{author}{\bibinfo{person}{Thomas Davidson}, \bibinfo{person}{Dana
  Warmsley}, \bibinfo{person}{Michael Macy}, {and} \bibinfo{person}{Ingmar
  Weber}.} \bibinfo{year}{2017}\natexlab{}.
\newblock \showarticletitle{Automated hate speech detection and the problem of
  offensive language}. In \bibinfo{booktitle}{\emph{Eleventh International AAAI
  Conference on Web and Social Media}}.
\newblock


\bibitem[Devlin et~al\mbox{.}(2018)]%
        {devlin2018bert}
\bibfield{author}{\bibinfo{person}{Jacob Devlin}, \bibinfo{person}{Ming-Wei
  Chang}, \bibinfo{person}{Kenton Lee}, {and} \bibinfo{person}{Kristina
  Toutanova}.} \bibinfo{year}{2018}\natexlab{}.
\newblock \showarticletitle{Bert: Pre-training of deep bidirectional
  transformers for language understanding}.
\newblock \bibinfo{journal}{\emph{arXiv preprint arXiv:1810.04805}}
  (\bibinfo{year}{2018}).
\newblock


\bibitem[Garg et~al\mbox{.}(2023)]%
        {Garg2023}
\bibfield{author}{\bibinfo{person}{Tanmay Garg}, \bibinfo{person}{Sarah Masud},
  \bibinfo{person}{Tharun Suresh}, {and} \bibinfo{person}{Tanmoy Chakraborty}.}
  \bibinfo{year}{2023}\natexlab{}.
\newblock \showarticletitle{{Handling Bias in Toxic Speech Detection: A
  Survey}}.
\newblock \bibinfo{journal}{\emph{Comput. Surveys}} (\bibinfo{year}{2023}).
\newblock
\showISBNx{2202.00126v3}
\showISSN{0360-0300}
\urldef\tempurl%
\url{https://doi.org/10.1145/3580494}
\showDOI{\tempurl}
\showeprint[arxiv]{2202.00126}


\bibitem[Grimminger and Klinger(2020)]%
        {grimminger-klinger-2020-hate}
\bibfield{author}{\bibinfo{person}{Lara Grimminger} {and}
  \bibinfo{person}{Roman Klinger}.} \bibinfo{year}{2020}\natexlab{}.
\newblock \showarticletitle{Hate Towards the Political Opponent: A Twitter
  Corpus Study of the 2020 \{US\} Elections on the Basis of Offensive Speech
  and Stance Detection}.
\newblock \bibinfo{journal}{\emph{Proceedings of the Eleventh Workshop on
  Computational Approaches to Subjectivity, Sentiment and Social Media
  Analysis}}, \bibinfo{pages}{187--197}.
\newblock
\urldef\tempurl%
\url{https://doi.org/10.18653/v1/2020.wassa-1.22}
\showDOI{\tempurl}


\bibitem[Gupta et~al\mbox{.}(2022)]%
        {gupta2022mitigating}
\bibfield{author}{\bibinfo{person}{Umang Gupta}, \bibinfo{person}{Jwala
  Dhamala}, \bibinfo{person}{Varun Kumar}, \bibinfo{person}{Apurv Verma},
  \bibinfo{person}{Yada Pruksachatkun}, \bibinfo{person}{Satyapriya Krishna},
  \bibinfo{person}{Rahul Gupta}, \bibinfo{person}{Kai-Wei Chang},
  \bibinfo{person}{Greg~Ver Steeg}, {and} \bibinfo{person}{Aram Galstyan}.}
  \bibinfo{year}{2022}\natexlab{}.
\newblock \showarticletitle{Mitigating gender bias in distilled language models
  via counterfactual role reversal}.
\newblock \bibinfo{journal}{\emph{arXiv preprint arXiv:2203.12574}}
  (\bibinfo{year}{2022}).
\newblock


\bibitem[Huang et~al\mbox{.}(2019)]%
        {huang2019reducing}
\bibfield{author}{\bibinfo{person}{Po-Sen Huang}, \bibinfo{person}{Huan Zhang},
  \bibinfo{person}{Ray Jiang}, \bibinfo{person}{Robert Stanforth},
  \bibinfo{person}{Johannes Welbl}, \bibinfo{person}{Jack Rae},
  \bibinfo{person}{Vishal Maini}, \bibinfo{person}{Dani Yogatama}, {and}
  \bibinfo{person}{Pushmeet Kohli}.} \bibinfo{year}{2019}\natexlab{}.
\newblock \showarticletitle{Reducing sentiment bias in language models via
  counterfactual evaluation}.
\newblock \bibinfo{journal}{\emph{arXiv preprint arXiv:1911.03064}}
  (\bibinfo{year}{2019}).
\newblock


\bibitem[Liu et~al\mbox{.}(2023)]%
        {liu2021pre}
\bibfield{author}{\bibinfo{person}{Pengfei Liu}, \bibinfo{person}{Weizhe Yuan},
  \bibinfo{person}{Jinlan Fu}, \bibinfo{person}{Zhengbao Jiang},
  \bibinfo{person}{Hiroaki Hayashi}, {and} \bibinfo{person}{Graham Neubig}.}
  \bibinfo{year}{2023}\natexlab{}.
\newblock \showarticletitle{{Pre-train, Prompt, and Predict: A Systematic
  Survey of Prompting Methods in Natural Language Processing}}.
\newblock \bibinfo{journal}{\emph{Comput. Surveys}} \bibinfo{volume}{55},
  \bibinfo{number}{9} (\bibinfo{year}{2023}).
\newblock
\showISSN{15577341}
\urldef\tempurl%
\url{https://doi.org/10.1145/3560815}
\showDOI{\tempurl}
\showeprint[arxiv]{2107.13586}


\bibitem[Liu et~al\mbox{.}(2021)]%
        {liu2021mitigating}
\bibfield{author}{\bibinfo{person}{Ruibo Liu}, \bibinfo{person}{Chenyan Jia},
  \bibinfo{person}{Jason Wei}, \bibinfo{person}{Guangxuan Xu},
  \bibinfo{person}{Lili Wang}, {and} \bibinfo{person}{Soroush Vosoughi}.}
  \bibinfo{year}{2021}\natexlab{}.
\newblock \showarticletitle{Mitigating political bias in language models
  through reinforced calibration}. In \bibinfo{booktitle}{\emph{Proceedings of
  the AAAI Conference on Artificial Intelligence}}, Vol.~\bibinfo{volume}{35}.
  \bibinfo{pages}{14857--14866}.
\newblock


\bibitem[May et~al\mbox{.}(2019)]%
        {may2019measuring}
\bibfield{author}{\bibinfo{person}{Emily May}, \bibinfo{person}{Yao Wang},
  \bibinfo{person}{Shikha Bordia}, \bibinfo{person}{Hannah Gao},
  \bibinfo{person}{Jilin Li}, {and} \bibinfo{person}{Percy Liang}.}
  \bibinfo{year}{2019}\natexlab{}.
\newblock \showarticletitle{Measuring and mitigating unintended bias in text
  classification}. In \bibinfo{booktitle}{\emph{Proceedings of the AAAI
  Conference on Artificial Intelligence}}, Vol.~\bibinfo{volume}{33}.
  \bibinfo{pages}{7479--7486}.
\newblock


\bibitem[Nadeem et~al\mbox{.}(2020)]%
        {nadeem2020stereoset}
\bibfield{author}{\bibinfo{person}{Moin Nadeem}, \bibinfo{person}{Anna Bethke},
  {and} \bibinfo{person}{Siva Reddy}.} \bibinfo{year}{2020}\natexlab{}.
\newblock \showarticletitle{StereoSet: Measuring stereotypical bias in
  pretrained language models}.
\newblock \bibinfo{journal}{\emph{arXiv preprint arXiv:2004.09456}}
  (\bibinfo{year}{2020}).
\newblock


\bibitem[Peng et~al\mbox{.}({[n.\,d.]})]%
        {gptwork}
\bibfield{author}{\bibinfo{person}{Baolin Peng}, \bibinfo{person}{Chunyuan Li},
  \bibinfo{person}{Pengcheng He}, \bibinfo{person}{Michel Galley}, {and}
  \bibinfo{person}{Jianfeng Gao}.} \bibinfo{year}{[n.\,d.]}\natexlab{}.
\newblock \showarticletitle{INSTRUCTION TUNING WITH GPT-4}.
\newblock  (\bibinfo{year}{[n.\,d.]}).
\newblock
\urldef\tempurl%
\url{https://instruction-tuning-with-gpt-4.github.io/}
\showURL{%
\tempurl}


\bibitem[Radford et~al\mbox{.}(2018)]%
        {radford2018improving}
\bibfield{author}{\bibinfo{person}{Alec Radford}, \bibinfo{person}{Karthik
  Narasimhan}, \bibinfo{person}{Tim Salimans}, {and} \bibinfo{person}{Ilya
  Sutskever}.} \bibinfo{year}{2018}\natexlab{}.
\newblock \showarticletitle{Improving language understanding with unsupervised
  learning}.
\newblock \bibinfo{journal}{\emph{Technical report, OpenAI}}
  (\bibinfo{year}{2018}).
\newblock


\bibitem[Raza et~al\mbox{.}(2022)]%
        {raza2022dbias}
\bibfield{author}{\bibinfo{person}{Shaina Raza}, \bibinfo{person}{Deepak~John
  Reji}, {and} \bibinfo{person}{Chen Ding}.} \bibinfo{year}{2022}\natexlab{}.
\newblock \showarticletitle{Dbias: detecting biases and ensuring fairness in
  news articles}.
\newblock \bibinfo{journal}{\emph{International Journal of Data Science and
  Analytics}} (\bibinfo{year}{2022}), \bibinfo{pages}{1--21}.
\newblock


\bibitem[Smith et~al\mbox{.}(2022)]%
        {smith-etal-2022-im}
\bibfield{author}{\bibinfo{person}{Eric~Michael Smith},
  \bibinfo{person}{Melissa Hall}, \bibinfo{person}{Melanie Kambadur},
  \bibinfo{person}{Eleonora Presani}, {and} \bibinfo{person}{Adina Williams}.}
  \bibinfo{year}{2022}\natexlab{}.
\newblock \showarticletitle{{``}{I}{'}m sorry to hear that{''}: Finding New
  Biases in Language Models with a Holistic Descriptor Dataset}. In
  \bibinfo{booktitle}{\emph{Proceedings of the 2022 Conference on Empirical
  Methods in Natural Language Processing}}. \bibinfo{publisher}{Association for
  Computational Linguistics}, \bibinfo{address}{Abu Dhabi, United Arab
  Emirates}, \bibinfo{pages}{9180--9211}.
\newblock
\urldef\tempurl%
\url{https://aclanthology.org/2022.emnlp-main.625}
\showURL{%
\tempurl}


\bibitem[Vig et~al\mbox{.}(2020)]%
        {vig2020investigating}
\bibfield{author}{\bibinfo{person}{Jesse Vig}, \bibinfo{person}{Sebastian
  Gehrmann}, \bibinfo{person}{Yonatan Belinkov}, \bibinfo{person}{Sharon Qian},
  \bibinfo{person}{Daniel Nevo}, \bibinfo{person}{Yaron Singer}, {and}
  \bibinfo{person}{Stuart Shieber}.} \bibinfo{year}{2020}\natexlab{}.
\newblock \showarticletitle{Investigating gender bias in language models using
  causal mediation analysis}.
\newblock \bibinfo{journal}{\emph{Advances in neural information processing
  systems}}  \bibinfo{volume}{33} (\bibinfo{year}{2020}),
  \bibinfo{pages}{12388--12401}.
\newblock


\bibitem[Wu et~al\mbox{.}(2022)]%
        {wu2022leveraging}
\bibfield{author}{\bibinfo{person}{Heting Wu}, \bibinfo{person}{Zhenxin Sun},
  \bibinfo{person}{Sheng Li}, {and} \bibinfo{person}{Xiang Ren}.}
  \bibinfo{year}{2022}\natexlab{}.
\newblock \showarticletitle{Leveraging Prompt Engineering and Counterfactual
  Data Augmentation for Fairer Language Models}. In
  \bibinfo{booktitle}{\emph{Proceedings of the 2022 Conference of the North
  American Chapter of the Association for Computational Linguistics
  (NAACL-HLT)}}.
\newblock


\end{thebibliography}

\end{document}